\documentclass[conference]{IEEEtran}
\IEEEoverridecommandlockouts
\usepackage{cite}
\usepackage{amsmath,amssymb,amsfonts}
\usepackage{algorithmic}
\usepackage{graphicx}
\usepackage{textcomp}
\usepackage{xcolor}
\usepackage{multirow}
\usepackage{marvosym}
\usepackage{hyperref}
\usepackage{cleveref}

\def\BibTeX{{\rm B\kern-.05em{\sc i\kern-.025em b}\kern-.08em
    T\kern-.1667em\lower.7ex\hbox{E}\kern-.125emX}}

\begin{document}

\title{Hierarchical Sub-action Tree  for Continuous Sign Language Recognition}

\author{
\IEEEauthorblockN{
Dejie Yang$^{1}$
Zhu Xu$^{1}$
Xinjie Gao$^{1}$
Yang Liu$^{1,2}$\textsuperscript{\Letter}}
\IEEEauthorblockA{$^{1}$Wangxuan Institute of  Computer Technology, Peking University, Beijing, China}
\IEEEauthorblockA{$^{2}$State Key Laboratory of General Artificial Intelligence, Peking Universitys, Beijing, China}

    \{ydj,xuzhu,gxj1914779162\}@stu.pku.edu.cn, yangliu@pku.edu.cn
\thanks{\textsuperscript{\Letter}Corresponding author.}
}

\maketitle

\begin{abstract}
Continuous sign language recognition (CSLR) aims to transcribe untrimmed videos into glosses, which are typically textual words. Recent studies indicate that the lack of large datasets and precise annotations has become a bottleneck for CSLR due to insufficient training data. To address this, some works have developed cross-modal solutions to align visual and textual modalities. However, they typically extract textual features from glosses without fully utilizing their knowledge. In this paper, we propose the Hierarchical Sub-action Tree (HST), termed HST-CSLR, to efficiently combine gloss knowledge with visual representation learning. By incorporating gloss-specific knowledge from large language models, our approach leverages textual information more effectively. Specifically, we construct an HST for textual information representation, aligning visual and textual modalities step-by-step and benefiting from the tree structure to reduce computational complexity. Additionally, we impose a contrastive alignment enhancement to bridge the gap between the two modalities. Experiments on four datasets (PHOENIX-2014, PHOENIX-2014T, CSL-Daily, and Sign Language Gesture) demonstrate the effectiveness of our HST-CSLR. Code and model are available at: \url{https://github.com/Federfallt/HST-CSLR.git}.

\end{abstract}


\section{Introduction}
\label{sec:intro}

Sign language serves as a vital communication tool for the deaf community. However, for hearing individuals, learning and mastering sign language can be challenging and time-intensive, creating barriers to direct communication between the two groups.  To address this issue, continuous sign language recognition (CSLR)  \cite{graves2006connectionist, zhou2020spatial, hao2021self, min2021visual, hu2023continuous} aims to translate consecutive frames  into a series of glosses\footnote{gloss is the atomic lexical unit in sign language, typically a word or phrase.} to express some sentences.

\begin{figure}[t]
\centering
\includegraphics[width=1\linewidth]{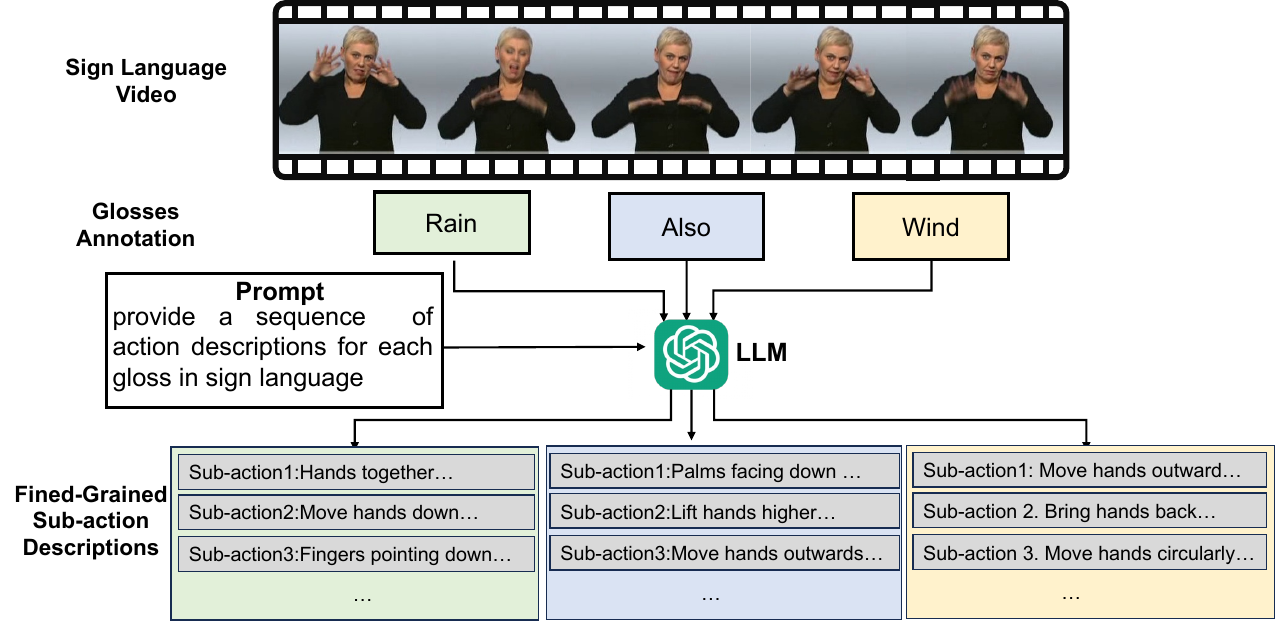}
\setlength{\abovecaptionskip}{-0.4cm} 
\caption{Sign language videos display a hierarchy of semantic information, from high-level events (glosses) to fine-grained subactions. However, existing datasets lack detailed annotations for these subactions, posing challenges for improving the layered understanding of sign language content. To address this, we leverage large language models (LLMs) to generate precise and meaningful descriptions of subactions, thereby enhancing the hierarchical understanding of sign language videos. }
\label{fig1}
\vspace{-0.4cm}
\end{figure}

Current CSLR models \cite{zhou2020spatial, hao2021self, min2021visual, hu2023continuous} typically include a spatial module (e.g., 2D-CNN) to extract short-term spatial features from the input clips, a temporal module (e.g., 1D-CNN or BiLSTM) to capture long-term dependencies, and a classifier that leverages the widely-used Connectionist Temporal Classification (CTC) loss \cite{graves2006connectionist} to predict the probability distribution of the target gloss sequence. 
Typically, current CSLR models can be categorized into two lines of approaches: The multi-stream CSLR framework \cite{chen2022two, zuo2022c2slr, jiao2023cosign} extends the spatial module with multi-cue information, such as key-points heatmap \cite{chen2022two, zuo2022c2slr} or skeleton information \cite{jiao2023cosign}, to guide model more focus on the hands and head information to acquire the most crucial semantic information.
{Recent approaches typically focus on visual features within sign language videos, and utilize cross-modal alignment module to align glosses and visual features, with the aim to improve gloss comprehension.} 
Compared to the complexity of the multi-stream framework, the cross-modal alignment framework is simple yet effective.
But how to better utilize the linguistic knowledge contained in glosses remains the key problem.
Though great successes, current methods do not model the complex temporal and contextual dependencies inherent in sign language video explicitly, restricting their ability, and how to better utilize such linguistic knowledge contained in glosses remains the key problem. As denoted in Fig.~\ref{fig1}, sign language videos present a hierarchical semantic structure, ranging from high-level events to fine-grained sub-actions, while two crucial reasons hinder its such hierarchical information exploration: (1) Existing datasets only encompass gloss-level annotations without supporting the fine-grained action-level understanding; (2) The movements in sign language have a clear temporal sequence, whose modeling is overlooked by previous approaches, and result in sub-optimal understanding of sign language.   
{To this end, we propose a novel approach termed Hierarchical Sub-action Tree for continuous sign language recognition (HST-CSLR). The key idea is to improve the understanding of sign language from temporal and contextual perspectives. Specifically, for the lack of fine-grained annotations within datasets, we propose to utilize Large Language Model (LLM) to explore the potential information for each gloss representation, acquiring detailed and sequential description for each gloss. Then to fully utilize these information while enable the temporal modeling of sign language, we introduce a hierarchical description tree using the obtained descriptions and design a Tree search algorithm, which not only integrate the knowledge of descriptions, but also takes the temporal information of each gloss into consideration, thus fully uncovering the comprehensive information stored within these descriptions to assist the sign understanding language in a cross-modal manner. Furthermore, to explicitly improve the visual and textual consistency, we impose a hierarchical  cross-modal alignment enhancement approach with the supervision of the  activated tree nodes. We evaluate our HST-CSLR on {four} CSLR benchmarks, including PHOENIX-2014 \cite{koller2015continuous}, PHOENIX-2014T \cite{camgoz2018neural}, CSL-Daily\cite{zhou2021improving} and Sign Language Gesture\cite{mohan2024classification}. Quantitative results indicate that we surpass existing state-of-the-art works. Through ablation study, we also verify the effectiveness of introducing modules.  }
Our main contributions are as follows: 
{(1) We develop a Hierarchical Description Tree search framework, which enables comprehensive contextual and temporal modeling for CSLR task; (2) We introduce novel alignment paradigms to enhance the alignment between visual and linguistic modalities of sign language, yielding more robust recognition performance. We also compensate fine-grained gloss descriptions within existing sign language datasets, facilitating further contextual understanding of continual sign language. (3) Experiments on four datasets demonstrate the effectiveness of our approach.}
\section{Related Work}
\label{sec:related}
\textbf{Continuous Sign Language Recognition}.
Continuous sign language recognition (CSLR) aims to translate sequences of image frames into corresponding glosses using a weakly supervised approach, where only sentence-level annotations are available as supervision.
And CSLR methods can be broadly categorized into two approaches: multi-cue and single-cue methods.
Muti-cue methods~\cite{ zuo2022c2slr, zhang2023c2st} incorporate multi-modal features including  hand shapes, facial expressions, mouths, and poses.
Such methods require extracting and integrating features from multiple modalities, resulting in high computational costs and making them challenging to apply in practical scenarios.
To address this issue, recent studies have focused on single-cue methods~\cite{zheng2023cvt,min2021visual}, which take a single cue—specifically, RGB frames—as input to directly decode glosses from video sequences.
VAC~\cite{min2021visual} considers the visual  encoder as the student and the textual gloss decoder as the teacher, utilizing knowledge distillation to  align the visual and textual modalities.
Similarly, 
CVT-SLR~\cite{zheng2023cvt} explores the pretrained knowledge of visual and textual modalities with a pretrained  variational autoencoder.
In this paper, we focus on  the single-cue setting for efficient training and inference.

However, existing methods fail to capture the hierarchical structure of temporal and semantic dependencies inherent in sign language videos, limiting the performance of continuous sign language recognition. To  explore the hierarchical structure of sign language videos and the fine-grained sub-action descriptions, we propose a hierarchical description tree building process that integrates the knowledge from descriptions while simultaneously accounting for the temporal information of each gloss. Additionally, to enhance the alignment between sign language videos and glosses, we introduce a contrastive alignment enhancement approach. This method explicitly improves visual and textual consistency and provides supervision for the nodes in the hierarchical tree.



\section{Proposed Method}
\begin{figure*}[t]
\centering
\vspace{-0.4cm}
\includegraphics[width=0.9\textwidth]{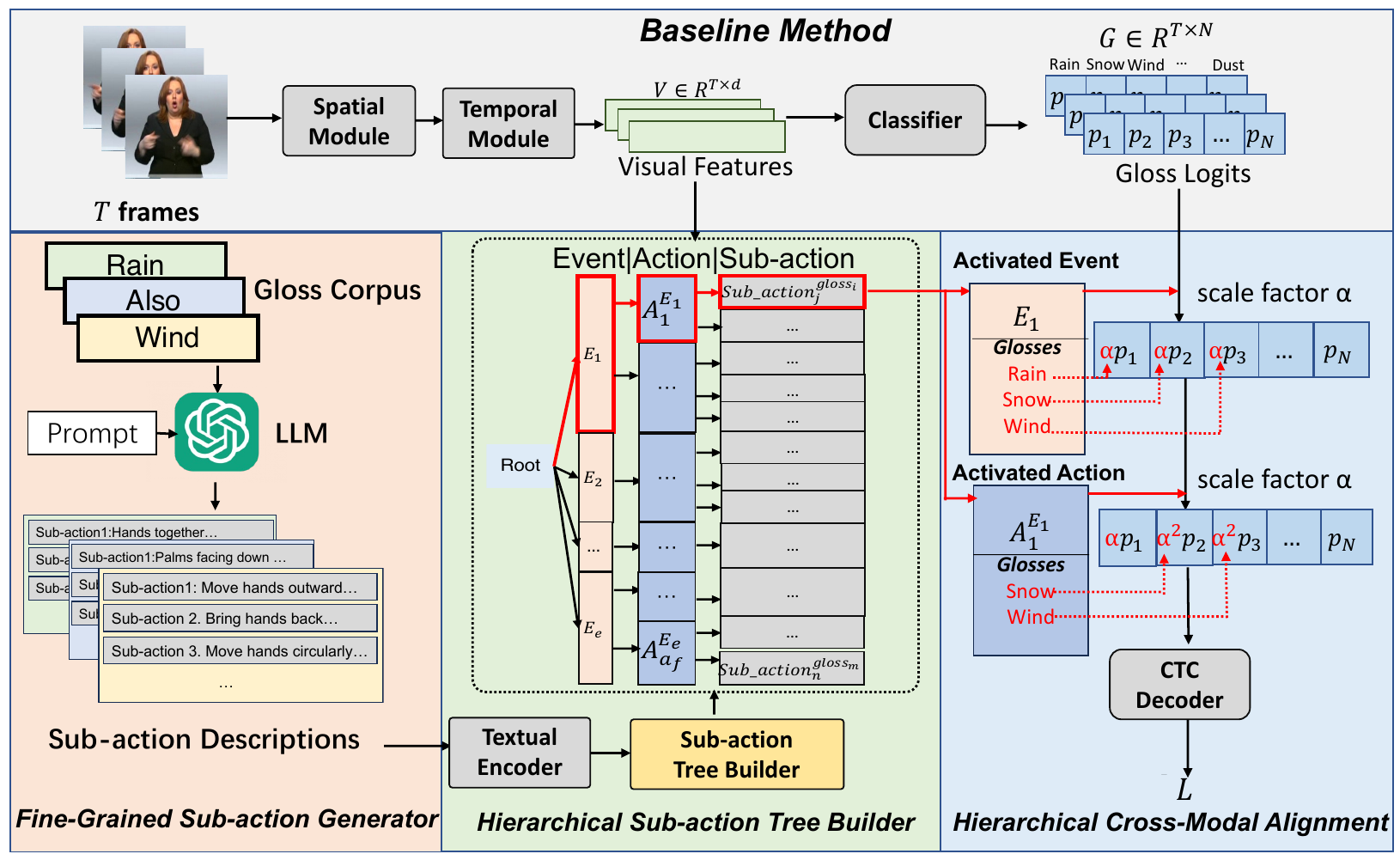}
\setlength{\abovecaptionskip}{-0.2cm} 
\caption{Our proposed HST-CSLR framework. To explore the  fine-grained sub-action information, we use an LLM to generate detailed descriptions for each gloss and construct a Hierarchical Sub-action Tree (HST). An optimal path search algorithm is applied to integrate semantic and temporal information in sub-action sequence, and a hierarchical cross-modal alignment enhances visual-textual consistency using activated tree nodes.}
\setlength{\belowcaptionskip}{-2cm}
\vspace{-0.5cm}
\label{fig2}
\end{figure*}

\subsection{Problem Definition}
Continuous Sign Language Recognition (CSLR) focuses on translating sequences of image frames into corresponding glosses. This task is typically approached in a weakly supervised setting, where only sentence-level annotations are provided as supervision, rather than frame-level or word-level labels.
Given a sign video with \textit{T} frames $X= \left\{ 
x_t \right\}^T_{t=1} \in \mathcal{R}^{T \times 3 \times H \times W}$, a CSLR model aims to translate it into a gloss sequence $Y= \left\{ 
y_i \right\}^N_{i=1}$, where $N$ denotes the length of the gloss sequence and is related to the video.

\subsection{Overview of Our Method}
As illustrated in \Cref{fig2}, our proposed  Hierarchical Sub-action Tree  for CSLR(HST-CSLR) framework contains four main components: a baseline 
 method, a fine-grained sub-action generator, a hierarchical sub-action tree builder and  a hierarchical cross-modal alignment module.

\subsubsection{Baseline method} The baseline consists  of a spatial module, a temporal module, a classifier. Specifically, to extract the spatil-temporal representation of sign video frames, the spatial module  processes the input frames into frame-level spatial features and  then the temporal module performs temporal modelling to obtain the  visual features $V\in \mathbb{R}^{T\times d}$.
And to calculate the probability sequence from the visual feature,  a classifier predicts the initial gloss logits for each frame. {The objective function of baseline consists of three loss terms, CTC loss $\mathcal{L}_{seq}$ for logits supervision, auxiliary loss $\mathcal{L}_{VE}$ to enhance classifier robustness and knowledge distillation loss $\mathcal{L}_{VA}$ for contextual information injection, which formulated as $\mathcal{L}_{base} = \mathcal{L}_{seq} + \mathcal{L}_{VE} + \mathcal{L}_{VA}$.}
However, the baseline method can only predict the glosses, limiting the model’s ability to learn detailed transitions and dependencies between the fine-grained sub-actions. 
Therefore, we introduced a hierarchical sub-action tree mechanism and utilized it to achieve multi-level cross-modal alignment between video and textual glosses.

\subsubsection{Fine-grained sub-action generator}
To address the lack of fine-grained annotations in existing datasets, we propose utilizing Large Language Models (LLMs) to explore the latent information within each gloss representation, generating detailed and sequential descriptions of each sub-action for every gloss.  Details provided in \Cref{sec:gen}.

\subsubsection{Hierarchical sub-action tree builder}
To fully leverage this information while enabling the temporal modeling of sign language, we design tailored prompts  to guide the LLM in generating sub-action descriptions relevant to the specific  gloss $y_i \in Y$, where i denotes $\rm i^{th}$ gloss.
These prompts are crafted to elicit detailed and context-aware descriptions of the associated sub-actions $S_i={Sub\_action_k^i}$
(where $k$ is the number of sub-actions) for the $i$-th gloss $y_i$, such as  gestures,  expressions, and posture. To further organize and make full use of these descriptions, we apply a hierarchical clustering method. This produces a hierarchical representation $E = \{E_1, E_2, \dots, E_e\}$, where $e$ represents the number of clustered groups at the top level. The second level contains the set of broader actions $A = \{A_1^{E_1}, A_2^{E_1}, \dots, A_{a_1}^{E_1},\dots, A_{a_e}^{E_e}\}$, where the superscripts$E_i$ shows the the second-level node belong to the parent node $E_i$, where ${a_i}$ is the is the number of mid-level actions for $E_i$, and the third level comprises the leaf nodes.
Details provided in \Cref{sec:tree}.

\subsubsection{Hierarchical cross-modal alignment}
To effectively leverage the path information within the Hierarchical Structure Tree (HST), we design a hierarchical cross-modal alignment module. This module is specifically crafted to fully exploit the hierarchical path information, enhancing the process of gloss recognition and improving overall model performance. We utilize visual features ${V}$ to select an optimal path in the Hierarchical Structure Tree (HST) by evaluating the matching degree at different hierarchical levels. Specifically, the optimal path ${P}^*$ is determined based on the alignment scores between ${V}$ and the nodes in the HST. Along the selected path ${P}^*$, we activate the nodes corresponding to the gloss $y^i$ and update the logits $\mathbf{p}^i$ of the gloss $y^i$ accordingly to refine the recognition process.
Details provided in \Cref{sec:align}.
 


\begin{figure}
    \centering
    \includegraphics[width=0.9\linewidth]{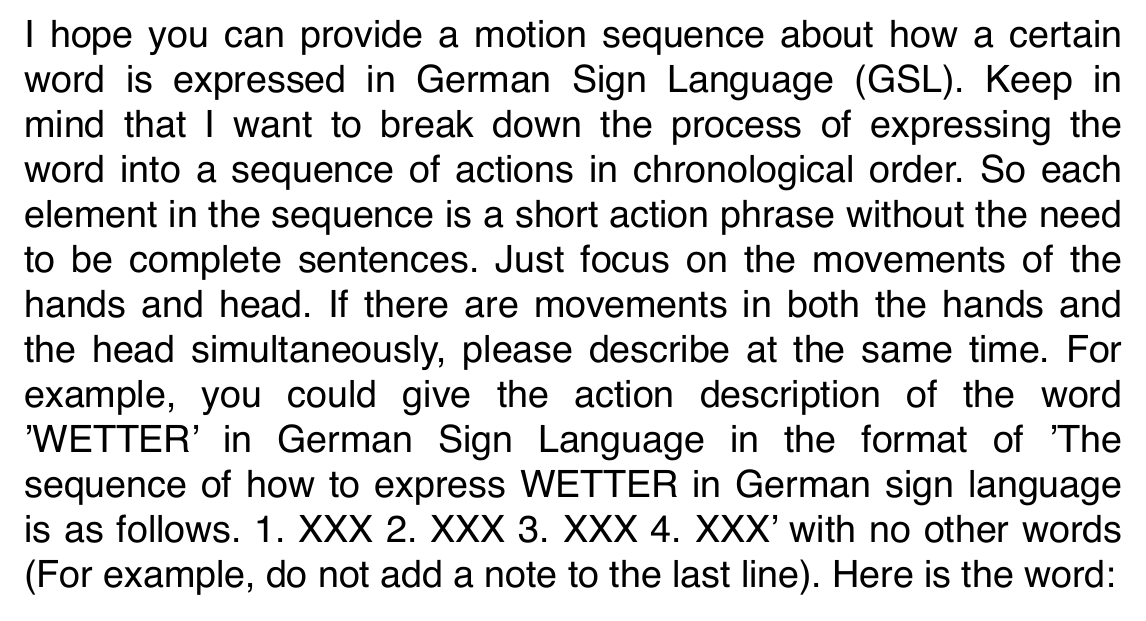}
    \setlength{\abovecaptionskip}{-0.2cm}
    \caption{An example of detailed prompts.}
    \label{fig_prompt}
    \setlength{\belowcaptionskip}{-0.8cm}
    \vspace{-0.6cm}
\end{figure}

\subsection{Fine-grained Sub-Action Generator}\label{sec:gen}
Effectively leveraging the semantic information embedded in glosses is crucial for enhancing model performance. We hypothesize that utilizing a Large Language Model (LLM) to generate textual descriptions of sign language actions can significantly improve the alignment between the visual and textual modalities, thereby boosting the model's overall performance. To ensure the generation of high-quality action descriptions, we employ GPT-4 as the description generator and meticulously craft prompts to optimize its output. The detailed prompts are shown in \Cref{fig_prompt}. As shown in \Cref{fig1}, guided by our carefully designed prompt,  LLM is capable of generating 4-5 detailed sub-action descriptions for each gloss, specifying aspects such as movements, motions, and postures.

Inevitably, some low-quality gloss statements, including incorrect or nonsensical words, arise in the generated sub-action descriptions. These flawed descriptions can have a noticeable adverse effect on the alignment process. To mitigate this issue, we propose manually filtering and regenerating such descriptions to eliminate their negative impact. 

\subsection{Hierarchical Sub-Action Tree Builder}\label{sec:tree}
To fully capture the hierarchical structure and semantic relationships within sign language glosses, we propose organizing sub-action descriptions into a tree-structured representation. This hierarchical modeling not only enables effective temporal alignment of visual features with textual descriptions but also allows for the progressive utilization of granular semantic information. By constructing a Hierarchical Sub-action Tree (HST) and performing optimal path searching, we aim to enhance cross-modal alignment and improve CSLR.

 \subsubsection{Hierarchical Clustering for HST Construction}
To organize the sub-action descriptions into a structured representation, we construct a three-level Hierarchical Sub-action Tree (HST). First, we apply K-means clustering on all sub-action descriptions based on their semantic similarity, creating $N_1$ first-level nodes $E = \{E_1, E_2, \dots, E_{N_1}\}$. Each $E_i$ contains sub-action descriptions with similar semantics.

To further refine the clusters, we perform a second round of clustering within each first-level node $E_i$ to form second-level nodes $A = \{A_1^{E_i}, A_2^{E_i}, \dots, A_{a_i}^{E_i}\}$, where $a_i$ is the number of second-level nodes under $E_i$. During this step, we ensure temporal consistency by including subsequent statements of the original clustering samples into the clustering set. Finally, each second-level node’s contents are treated as individual leaf nodes, forming the third layer of the HST. This hierarchical clustering process organizes sub-actions into semantically and temporally coherent groups, resulting in a  three-level tree.

\subsubsection{Optimal Path Searching for Visual-Action Matching}
To align visual features with the HST, we design an optimal path search algorithm. Let the visual feature be $V \in \mathbb{R}^{T \times d}$, which we compare with the textual features $t^E_i \in \mathbb{R}^{1 \times d}$, extracted from the each node $E_i$ of first-level nodes using a text encoder. The cosine similarity is calculated as:

\begin{equation}
    d_{V, t^E_i} = \cos(V, t^E_i),
\end{equation}
\begin{equation}
    p^E_i = \mathop{\mathrm{argmax}} d_{V, t^E_i}.
\end{equation}

The node $p^E_i$ with the highest similarity is selected, and the process is similarly repeated for the second-level nodes $A^{E_i}_j$  ( the children nodes of $E_i$) and their textual features $t^A_j \in \mathbb{R}^{1 \times d}$. Finally, this search is extended to the leaf nodes (third layer), where the textual feature of each sub-action is matched with $V$ to find the best alignment at the most granular level. The resulting path $p = \{p^E_i, p^A_j, p^S_k\}$ connects the root to the best-matching leaf node.


\subsection{Hierarchical Cross-Modal Alignment}\label{sec:align}

\subsubsection{Hierarchical Updating}
How to utilize the path information within HST is crucial to enhance the visual and textual alignment. We design an hierarchical updating approach to fully use the path information to assist gloss recognition.
Specifically, as each node on the path contains a series of statements, we record the gloss index that these statements belong to connect fine-grained statements with glosses. Then we highlight the corresponding logits within the classifier output $l_{origin} \in \mathcal{R}^{T \times N}$, where $N$ is the number of classes, by multiplying an update scale $\alpha$. Notably, the updating degree is decided by the occurance within the hierarchical path. Thus, for the gloss corresponding to the statement in leaf node, it should have been updated three times, i.e., multiplied by $\alpha^{3}$. The rationale behind this operation is that, the gloss that appears multiple times on the path clearly corresponds to the video clip to a higher degree, and should according acquire higher concentration. By implementing the hierarchical updating approach, we fully exploit the stored information with path, gradually strengthening the consistency between vision and text, then refining the recognition process. In practice, we use an update matrix $W \in \mathcal{R}^{T \times N}$ to record the update coefficients and calculate Hadamard product between $W$ and $l_{origin}$ to obtain the updated logits $l_{updated}$:

\begin{equation}
    l_{updated} = l_{origin} \odot W.
\end{equation}

\subsubsection{Cross-Modal Contrastive Alignment}
Whether the model can find the correct hierarchical path is the key to the effectiveness of our HST building and updating. However, the path may introduce extra noise and wrong gloss information without additional supervision, which is originated from the low quality alignment between visual and textual modality. To improve the alignment quality, we further introduce a contrastive alignment loss. Firstly, we adopt the argmax within logits $l_{origin}$ to generate the pseudo label $y^*$ for a single gloss. Then we select the corresponding nodes in each layer matches the pseudo labels, i.e., contain statements corresponding to $y^*$, as positive samples, and other nodes as negative samples. The contrastive loss for each layer is formulated as follows:
\begin{equation}
    \mathcal{L}_{c} = - \frac{1}{N_{positive}} \log \frac{\sum_{i \in p_{positive}} \exp (cos(v, t_{k,i}))}{\sum_{j \in p_{negtive}} \exp (cos(v, t_{k,j}))},
\end{equation}
where $p_{positive}$ is the set of positive samples, $p_{negtive}$ is the set of negative samples, $k$ is the layer number, and $N_{positive}$ is the number of $p_{positive}$.


\subsection{Training and  Inference}
\subsubsection{Training:} The Objective Function is formulated as follow, $\mathcal{L} = \mathcal{L}_{base} + \mathcal{L}_{c}$, where $\mathcal{L}_{base}$ is the objective our baseline adopted, and $\mathcal{L}_{c}$ is the contrast alignment enhancement item we proposed.

\subsubsection{Inference:}
During inference, we first utilize the baseline model to extract the visual feature and predict an initial logits  for all glosses. We then search for the optimal path  in the Hierarchical Sub-action Tree (HST) based on the alignment between visual feature and the nodes at each tree level. Using the selected path, we update the corresponding logits  to refine the predictions. Finally, the updated logits  are used as the final prediction result for gloss recognition.


\section{Experiments}
\subsection{Implementation Details}
Following previous approaches\cite{rao2024cross,lu2024tcnet}, we use four datasets: PHOENIX-2014 \cite{koller2015continuous}, PHOENIX-2014T \cite{camgoz2018neural}, CSL-Daily\cite{zhou2021improving} and Sign Language Gesture\cite{mohan2024classification}.
For evaluation metric, we also follow previous work CorrNet\cite{hu2023continuous}, which is our baseline, to adopt word error rate (WER). 
 For a fair comparison, we follow the structure of previous work CorrNet\cite{hu2023continuous}, which adopts ResNet-18 as backbone. To encode the descriptions, we use the BERT \cite{devlin2018bert} as the text encoder. We train our models for 80 epochs with initial learning rate 0.001 which is divided by 5 at epoch 40 and 60. Adam optimizer is adopted with weight decay 0.001 and batch size 2.  We set update scale $\alpha$ in Hierarchical Updating as 1.5.


\begin{table}[!ht]
\vspace{-0.3cm}
\setlength{\abovecaptionskip}{-0.1cm} 
\caption{Comparisons on PHOENIX-2014 \cite{koller2015continuous}, PHOENIX-2014T \cite{camgoz2018neural}.}

\centering
\begin{tabular}{c|cc|cc}
\hline
\multirow{2}{*}{Model} & \multicolumn{2}{c|}{PHOENIX-2014} & \multicolumn{2}{c}{PHOENIX-2014T}\\ 
& Dev $\downarrow$ & Test $\downarrow$ & Dev $\downarrow$ & Test $\downarrow$ \\
\hline
SubUNets \cite{cihan2017subunets} & 40.8 & 40.7 & - & - \\
Staged-Opt \cite{cui2017recurrent} & 39.4 & 38.7 & - & -\\
Align-iOpt \cite{pu2019iterative} & 37.1 & 36.7 & - & - \\
SFL \cite{niu2020stochastic} & 26.2 & 26.8 & 25.1 & 26.1 \\
VAC \cite{min2021visual} & 21.2 & 22.3 & - & - \\
STMC \cite{zhou2020spatial} & 21.1 & 20.7 & 19.6 & 21.0 \\
SMKD \cite{hao2021self} & 20.8 & 21.0 & 20.8 & 22.4 \\
$\mathrm{C}^2$SLR \cite{zuo2022c2slr} & 20.5 & 20.4 & 20.2 & 20.4 \\
CVT-SLR \cite{zheng2023cvt} & 19.8 & 20.1 & 19.4 & 20.3 \\
CorrNet \cite{hu2023continuous} & 18.8 & 19.4 & 18.9 & 20.5 \\
Two-Stream SLR \cite{chen2022two} & 18.4 & 18.8 & 17.7 & 19.3 \\
TCNet \cite{lu2024tcnet} & 18.1 & 18.9 & 18.3 & 19.4 \\
CSGC \cite{rao2024cross} & 18.1 & 19.0 & \textbf{17.2} & 19.5 \\
\hline
Ours & \textbf{17.9} & \textbf{18.2} & 17.4 & \textbf{19.1} \\
\hline
\end{tabular}
\label{table1}
\vspace{-0.3cm}
\end{table}

\subsection{Comparison with State-of-the-arts}

We compare our result with existing state-of-the-arts approaches, the results are shown in \Cref{table1,tab:csl-daily,tab:slg} . 
Experiments on the four datasets  show our generalization  across German, Chinese, and English, performing effectively on both videos and images.
Credit to our hierarchical updating process, we harvest performance gain and outperform previous state-of-the-art approaches, such as 0.2$\%$/0.6$\%$ performance gain on the Dev/Test set of PHOENIX-2014 \cite{koller2015continuous}, and 0.2\% on the Test set of PHOENIX-2014T \cite{camgoz2018neural}, which indicates that our fine-grained hierarchical Sub-action Tree and updating design can effectively provide comprehensive contextual and temporal modeling information.
However, the improvement on the Dev set of PHOENIX-2014T \cite{camgoz2018neural}  is limited. We attribute it to the inherent difficulty of acquiring high-quality yet fine-grained descriptions within these datasets as they contain noised and obscure glosses. We leave it as a further researching direction to provide more accurate fine-grained descriptions to enhance the performance.
\begin{table}[!ht]
    \centering
    \vspace{-0.5cm}
    \setlength{\abovecaptionskip}{-0.1cm} 
    \caption{Comparisons on CSL-Daily reported with WER(\%).}
\setlength{\abovecaptionskip}{0cm} 
\setlength{\belowcaptionskip}{-2cm}
    \begin{tabular}{ccccc}
        \hline
        & SEN\cite{hu2023self} & CorrNet\cite{hu2023continuous} & TCNet\cite{lu2024tcnet} & Ours \\
        \hline
        Dev & 31.1 & 30.6 & 29.7 & \textbf{27.5} \\
        Test & 30.7 & 30.1 & 29.3 & \textbf{27.4} \\
        \hline
    \end{tabular}
    \label{tab:csl-daily}
    \vspace{-0.6cm}
\end{table}
\begin{table}[!ht]
    \centering
    \vspace{-0.1cm}
    \setlength{\abovecaptionskip}{-0.1cm} 
    \caption{Comparisons on Sign Language Gesture\cite{mohan2024classification}.}
    \label{tab:slg}
    \begin{tabular}{cc}
    \hline
    Model & Accuracy ($\%$) \\
       \hline
        Baseline & 97.28 \\
        Ours & \textbf{99.85} \\
\hline
\end{tabular}
\vspace{-0.2cm}
\end{table}


\subsection{Ablation Study}

{All ablation studies are conducted on PHOENIX-2014T \cite{camgoz2018neural} dataset and reported with WER(\%).}

\begin{table}[!ht]
\vspace{-0.3cm}
\setlength{\abovecaptionskip}{0cm} 
\setlength{\belowcaptionskip}{-2cm}
\caption{Performance with different inputs. }
\centering
\begin{tabular}{c|c|c c}
\hline
\multirow{2}{*}{$\mathcal{L}_{\mathrm{Seq}}$ calculate} & \multirow{2}{*}{Decoder Input} & \multicolumn{2}{c}{WER ($\%$)} \\
& & Dev & Test\\
\hline
\multirow{2}{*}{w/ $l_{origin}$} & $l_{origin}$ & 18.9 & 20.5\\
& $l_{updated}$ & - & -\\
\hline
\multirow{2}{*}{w/ $l_{updated}$} & $l_{origin}$ & 18.2 & 19.4\\
& $l_{updated}$ & \textbf{17.4} & \textbf{19.0}\\
\hline
\end{tabular}
\label{table2}
\vspace{-0.6cm}
\end{table}
\noindent\textbf{Impacts of Hierarchical Updating:} We separately use origin logits $l_{origin}$ and our updated logits $l_{updated}$ as inputs for the CTC loss $\mathcal{L}_{\mathrm{Seq}}$ during training, and $l_{origin}$ and $l_{updated}$ as inputs for the CTC decoder to calculate WER. As illustrated in Table \ref{table2}, comparing with $l_{origin}$, $l_{updated}$ demonstrate 0.8\%/0.4\% performance improvement for Dev/Test split, no matter it serves as $\mathcal{L}_{\mathrm{Seq}}$ input during training or as the CTC decoder input during testing. 
\begin{table}[!ht]
\vspace{-0.4cm}
\setlength{\abovecaptionskip}{-0.1cm} 
\setlength{\belowcaptionskip}{-2cm}
\caption{Comparisons between different updating scale $\alpha$ on Dev set.}
\centering
\begin{tabular}{c|cccc}
\hline
$\alpha$ & 1.25 & 1.5 & 1.75 & 2.0 \\
\hline
WER ($\%$) & 18.4 & \textbf{17.4} & 18.6 & 19.5\\
\hline
\end{tabular}
\label{table3}
\vspace{-0.2cm}
\end{table}

\noindent\textbf{Impacts of Updating Scale $\alpha$:} We study the impact of different updating scale on performance, which is shown in Table \ref{table3}. Serving $\alpha$ as 1.5 acquires the best performance of 17.4 on Dev Set of PHOENIX-2014T. Too small $\alpha$ is less influential on the prediction process and fail to exert significant impact upon the performance, while too large $\alpha$ may hinder the robustness of the model and result in a decrease in performance.
\begin{table}[!ht]
\vspace{-0.6cm}
   \setlength{\abovecaptionskip}{-0.1cm}
\caption{Ablation of the tree depth(cluster parameters).}
\centering
\begin{tabular}{c|c c c c}
\hline
 & CorrNet(baseline) & 1 & 2 & 3 \\
 \hline
Dev & 18.9 & 18.2 & 17.8 & \textbf{17.4} \\
Test & 20.5 & 20.1 & 19.5 & \textbf{19.0} \\
\hline
\end{tabular}
\vspace{-0.2cm}
\label{table5}
\end{table}

\noindent\textbf{Impacts of Tree Structure:} We analyze the influence of different tree structures, which mainly concentrates on the tree depth. The performance is shown in Table.~\ref{table5}. As shown, adopting tree structure with one or two layers is insufficient to provide fine-grained visual and temporal information for performance improvement. With three layers to build the sub-action tree, we harvest 1.5\%/1.5\% performance gain on the Dev/Test, indicating that fine-grained sub-actions can explicitly provide valuable information for the gloss recognition.

\begin{table}[!h]
    \centering
    \vspace{-0.3cm}
    \setlength{\abovecaptionskip}{-0cm} 
    \caption{Positive-Negative Samples}
    \label{tab:pns}
    \begin{tabular}{ccc}
    \hline
       Model & Dev & Test \\
        \hline
        w/o & 17.5 & 19.6 \\
        w/ & \textbf{17.4} & \textbf{19.0} \\
        \hline
\end{tabular}
\vspace{-0.3cm}
\end{table}
\noindent\textbf{Impacts of Positive-Negative Samples:} Nodes in the tree that contain description sentences corresponding to this pseudo-label are treated as \textbf{positive}(pos), while the remaining nodes are considered \textbf{negative}(neg). As shown in \Cref{tab:pns}, we can improve the performance  with pos-neg samples.

\begin{table}[!ht]
\vspace{-0.2cm}
   \setlength{\abovecaptionskip}{-0.1cm}
\caption{Ablation on the HST and Cross-modal Alignment.}
\centering
\begin{tabular}{c|c c c c}
\hline
 & CorrNet(baseline) & baseline + HST & baseline + $\mathcal{L}_{c}$ & our \\
\hline
Dev & 18.9 & 17.6 & 18.1 & \textbf{17.4} \\
Test & 20.5 & 19.6 & 19.8 & \textbf{19.0} \\
\hline
\end{tabular}
\vspace{-0.4cm}
\label{table4}
\end{table}
\noindent\textbf{Impacts of Cross-modal Alignment:} The influence of Cross-model Alignment is shown in ~\Cref{table4}. As shown, by applying Hierarchical Sub-action Tree and Updating, we harvest 0.7\%/0.4\% performance gain in comparison with baseline. Further introducing cross-modal alignment upon it can receive 0.2\%/0.6\% further performance gain, indicating that introducing such cross-modal alignment can effectively well-aligned the visual and textual feature, leading to enhanced gloss recognition ability.

\begin{table}[!h]
    \centering
    \vspace{-0.5cm}
    \setlength{\abovecaptionskip}{-0.1cm} 
    \caption{Filtering methods}
    \label{tab:filter}
    \begin{tabular}{*{10}{c}}
        \hline
       Method & Dev & Test \\
        \hline
        LLMs & 17.9 & 19.6 \\
        Manual & 17.4 & 19.0 \\
        \hline
    \end{tabular}
    \vspace{-0.3cm}
\end{table}
\noindent\textbf{Impacts of Manual Filtering:} As shown in \Cref{tab:filter}, manual filtering outperforms LLM self-filtering, but the overall performance of LLM self-filtering still surpasses that of most existing methods.

\begin{figure}[!h]
\vspace{-0.4cm}
\includegraphics[width=\linewidth]{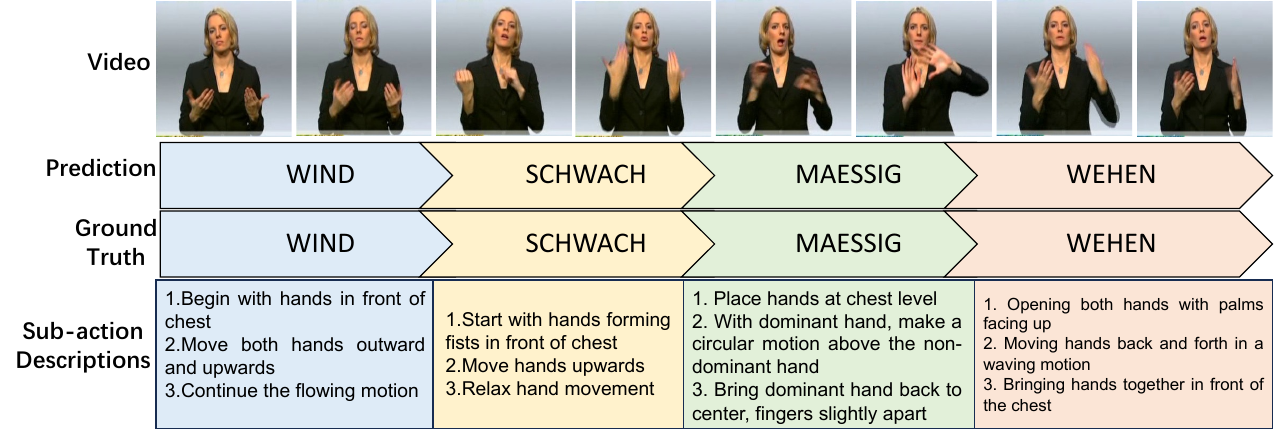}
\setlength{\abovecaptionskip}{-0.5cm} 
   \caption{Visual example.} 
    \label{fig:img1}
    \vspace{-0.2cm}
\end{figure}
\noindent\textbf{Visual examples:}
\Cref{fig:img1} shows a visual example, including sub-action descriptions from the LLM and comparison between our prediction and ground truth. The consistency between our prediction and ground truth shows the effectiveness of our method.

\section{Conclusion}
In this paper, to address the lack of fine-grained understanding pf continuous sign Language video, we propose the Hierarchical Sub-action Tree for CSLR (HST-CSLR), which efficiently integrates gloss-specific knowledge derived from large language models (LLMs) with visual representation learning. By constructing a Hierarchical Sub-action Tree (HST), our approach aligns visual and textual modalities while leveraging structured representation to reduce complexity. Furthermore, we introduce contrastive alignment enhancement to better bridge the cross-modal gap. Experiments demonstrate that HST-CSLR achieves state-of-the-art or competitive performance, showing its effectiveness in improving CSLR.
\section{Acknowledgment}
This work was supported by the grants from the National Natural Science Foundation of China 62372014 and Beijing Natural Science Foundation 4252040.

\bibliographystyle{IEEEbib}
\bibliography{main}

\begin{thebibliography}{10}

\bibitem{graves2006connectionist}
Alex Graves, Santiago Fern{\'a}ndez, Faustino Gomez, and J{\"u}rgen Schmidhuber,
\newblock ``Connectionist temporal classification: labelling unsegmented sequence data with recurrent neural networks,''
\newblock in {\em Proceedings of the 23rd international conference on Machine learning}, 2006.

\bibitem{zhou2020spatial}
Hao Zhou, Wengang Zhou, Yun Zhou, and Houqiang Li,
\newblock ``Spatial-temporal multi-cue network for continuous sign language recognition,''
\newblock in {\em Proceedings of the AAAI conference on artificial intelligence}, 2020.

\bibitem{hao2021self}
Aiming Hao, Yuecong Min, and Xilin Chen,
\newblock ``Self-mutual distillation learning for continuous sign language recognition,''
\newblock in {\em Proceedings of the IEEE/CVF international conference on computer vision}, 2021.

\bibitem{min2021visual}
Yuecong Min, Aiming Hao, Xiujuan Chai, and Xilin Chen,
\newblock ``Visual alignment constraint for continuous sign language recognition,''
\newblock in {\em Proceedings of the IEEE/CVF international conference on computer vision}, 2021.

\bibitem{hu2023continuous}
Lianyu Hu, Liqing Gao, Zekang Liu, and Wei Feng,
\newblock ``Continuous sign language recognition with correlation network,''
\newblock in {\em Proceedings of the IEEE/CVF Conference on Computer Vision and Pattern Recognition}, 2023.

\bibitem{chen2022two}
Yutong Chen, Ronglai Zuo, Fangyun Wei, Yu~Wu, Shujie Liu, and Brian Mak,
\newblock ``Two-stream network for sign language recognition and translation,''
\newblock {\em Advances in Neural Information Processing Systems}, 2022.

\bibitem{zuo2022c2slr}
Ronglai Zuo and Brian Mak,
\newblock ``C2slr: Consistency-enhanced continuous sign language recognition,''
\newblock in {\em Proceedings of the IEEE/CVF Conference on Computer Vision and Pattern Recognition}, 2022.

\bibitem{jiao2023cosign}
Peiqi Jiao, Yuecong Min, Yanan Li, Xiaotao Wang, Lei Lei, and Xilin Chen,
\newblock ``Cosign: Exploring co-occurrence signals in skeleton-based continuous sign language recognition,''
\newblock in {\em Proceedings of the IEEE/CVF International Conference on Computer Vision}, 2023.

\bibitem{koller2015continuous}
Oscar Koller, Jens Forster, and Hermann Ney,
\newblock ``Continuous sign language recognition: Towards large vocabulary statistical recognition systems handling multiple signers,''
\newblock {\em Computer Vision and Image Understanding}, 2015.

\bibitem{camgoz2018neural}
Necati~Cihan Camgoz, Simon Hadfield, Oscar Koller, Hermann Ney, and Richard Bowden,
\newblock ``Neural sign language translation,''
\newblock in {\em Proceedings of the IEEE conference on computer vision and pattern recognition}, 2018.

\bibitem{zhou2021improving}
Hao Zhou, Wengang Zhou, Weizhen Qi, Junfu Pu, and Houqiang Li,
\newblock ``Improving sign language translation with monolingual data by sign back-translation,''
\newblock in {\em Proceedings of the IEEE/CVF Conference on Computer Vision and Pattern Recognition (CVPR)}, 2021.

\bibitem{mohan2024classification}
Aadershi Mohan, Daivik Mohan, Satvik Vats, Vikrant Sharma, and Vinay Kukreja,
\newblock ``Classification of sign language gestures using cnn with adam optimizer,''
\newblock in {\em 2024 2nd International Conference on Disruptive Technologies (ICDT)}. IEEE, 2024.

\bibitem{zhang2023c2st}
Huaiwen Zhang, Zihang Guo, Yang Yang, Xin Liu, and De~Hu,
\newblock ``C2st: Cross-modal contextualized sequence transduction for continuous sign language recognition,''
\newblock in {\em Proceedings of the IEEE/CVF International Conference on Computer Vision}, 2023.

\bibitem{zheng2023cvt}
Jiangbin Zheng, Yile Wang, Cheng Tan, Siyuan Li, Ge~Wang, Jun Xia, Yidong Chen, and Stan~Z Li,
\newblock ``Cvt-slr: Contrastive visual-textual transformation for sign language recognition with variational alignment,''
\newblock in {\em Proceedings of the IEEE/CVF conference on computer vision and pattern recognition}, 2023.

\bibitem{rao2024cross}
Qi~Rao, Ke~Sun, Xiaohan Wang, Qi~Wang, and Bang Zhang,
\newblock ``Cross-sentence gloss consistency for continuous sign language recognition,''
\newblock in {\em Proceedings of the AAAI Conference on Artificial Intelligence}, 2024.

\bibitem{lu2024tcnet}
Hui Lu, Albert~Ali Salah, and Ronald Poppe,
\newblock ``Tcnet: Continuous sign language recognition from trajectories and correlated regions,''
\newblock in {\em Proceedings of the AAAI Conference on Artificial Intelligence}, 2024, vol.~38.

\bibitem{devlin2018bert}
Jacob Devlin,
\newblock ``Bert: Pre-training of deep bidirectional transformers for language understanding,''
\newblock {\em arXiv preprint arXiv:1810.04805}, 2018.

\bibitem{cihan2017subunets}
Necati Cihan~Camgoz, Simon Hadfield, Oscar Koller, and Richard Bowden,
\newblock ``Subunets: End-to-end hand shape and continuous sign language recognition,''
\newblock in {\em Proceedings of the IEEE international conference on computer vision}, 2017.

\bibitem{cui2017recurrent}
Runpeng Cui, Hu~Liu, and Changshui Zhang,
\newblock ``Recurrent convolutional neural networks for continuous sign language recognition by staged optimization,''
\newblock in {\em Proceedings of the IEEE conference on computer vision and pattern recognition}, 2017.

\bibitem{pu2019iterative}
Junfu Pu, Wengang Zhou, and Houqiang Li,
\newblock ``Iterative alignment network for continuous sign language recognition,''
\newblock in {\em Proceedings of the IEEE/CVF conference on computer vision and pattern recognition}, 2019.

\bibitem{niu2020stochastic}
Zhe Niu and Brian Mak,
\newblock ``Stochastic fine-grained labeling of multi-state sign glosses for continuous sign language recognition,''
\newblock in {\em Computer Vision--ECCV 2020: 16th European Conference, Glasgow, UK, August 23--28, 2020, Proceedings, Part XVI 16}. Springer, 2020.

\bibitem{hu2023self}
Lianyu Hu, Liqing Gao, Zekang Liu, and Wei Feng,
\newblock ``Self-emphasizing network for continuous sign language recognition,''
\newblock in {\em Thirty-seventh AAAI conference on artificial intelligence}, 2023.

\end{thebibliography}
\end{document}